\documentclass[10pt]{article}

\usepackage{amsmath}
\usepackage{amssymb}
\usepackage{array}
\usepackage{booktabs}
\usepackage{caption}
\usepackage{framed}
\usepackage[margin=0.76in]{geometry}
\usepackage{graphicx}
\usepackage[round,authoryear]{natbib}
\usepackage[section]{placeins}
\usepackage{tabularx}
\usepackage{titlesec}
\usepackage[table]{xcolor}
\usepackage{hyperref}

\definecolor{noteblue}{HTML}{234A73}
\definecolor{noteink}{HTML}{18212B}
\definecolor{notepanel}{HTML}{F1F4F7}
\definecolor{noterule}{HTML}{C7D0D9}
\definecolor{shadecolor}{HTML}{F1F4F7}

\hypersetup{
  colorlinks = true,
  linkcolor  = {noteblue},
  citecolor  = {noteblue},
  urlcolor   = {noteblue},
  pdftitle   = {Stabilized Best-of-K Training for Neural Combinatorial Optimization},
  pdfauthor  = {Melveena Jolly and Midhun Xavier}
}
\titleformat{\section}
  {\large\bfseries\color{noteink}}
  {}{0pt}{}
  [\vspace{0.12em}\color{noterule}\titlerule]
\titleformat{\subsection}
  {\normalsize\bfseries\color{noteink}}
  {}{0pt}{}
\titlespacing*{\section}{0pt}{1.7ex plus 0.5ex minus 0.2ex}{0.8ex}
\titlespacing*{\subsection}{0pt}{1.2ex plus 0.3ex minus 0.2ex}{0.35ex}

\newcolumntype{Y}{>{\raggedright\arraybackslash}X}
\newcommand{\E}{\mathbb{E}}
\newcommand{\notelabel}[1]{%
  {\small\bfseries\color{noteblue}\MakeUppercase{#1}}%
}
\newenvironment{summarybox}{%
  \begin{snugshade}\noindent
}{%
  \end{snugshade}
}

\begin{document}

\begin{center}
  \notelabel{Experimental Note}\quad
  {\color{noterule}\textbullet}\quad
  {\small July 2026}

  \vspace{0.65em}
  {\LARGE\bfseries\color{noteink}
  Stabilized Best-of-$K$ Training for Neural Combinatorial Optimization\par}

  \vspace{0.65em}
  {\normalsize
  Melveena Jolly\textsuperscript{1}\quad
  Midhun Xavier\textsuperscript{1}\par}
  \vspace{0.15em}
  {\small
  \textsuperscript{1}Independent Researcher\quad
  \href{mailto:melveenajollyk@gmail.com}{melveenajollyk@gmail.com}\quad
  \href{mailto:midhunxavier@outlook.com}{midhunxavier@outlook.com}}
\end{center}

\begin{summarybox}
\notelabel{Summary}

\vspace{0.35em}
Leader Reward modifies POMO training to emphasize the best trajectory produced
by repeated inference. We test a narrow extension: replace its binary
leader/non-leader distinction with a stabilized rank signal indexed by a
sampling budget $K$. With the POMO architecture, 3,050-epoch schedule, and
TSP-100 test set held fixed, the Leader Reward reimplementation obtains
$7.7662$ under 100-start, 8-augmentation greedy decoding, matching the reported
$7.766$ at its displayed precision. Under independent sampling, the stabilized
$K=8$ recipe lowers realized Best-of-8 cost in all three paired training seeds:
$7.7944$ versus $7.8136$. This observation is estimation-only and
decoder-specific: three seeds are below the six-seed testing floor, Leader
Reward is better at sampled $K=1$, and it remains slightly better under its
original augmented-greedy protocol. We make no unbiased-estimator, universal
superiority, or state-of-the-art claim.
\end{summarybox}

\section{The question}

POMO solves an instance from multiple starting nodes and trains the resulting
trajectories with a shared-baseline policy gradient \citep{kwon2020pomo}.
Leader Reward \citep{wang2024leaderreward} observes that deployment retains the
best solution found within an inference budget, whereas the standard POMO
update values the entire multi-start batch. It increases the current leader's
weight during main training and uses leader-only fine-tuning.

We ask whether training should distinguish not only the current leader, but
also trajectories that can become the best member of a size-$K$ group when
deployment draws exactly $K$ independent trajectories. The motivating target is
\begin{equation}
  J_K(\theta)
  =
  \E_{\tau_1,\ldots,\tau_K\stackrel{\mathrm{iid}}{\sim}\pi_\theta}
  \left[\max_{1\leq j\leq K}R(\tau_j)\right].
  \label{eq:bestofk}
\end{equation}
The experiment changes only the trajectory-weight rule of the same POMO
policy. It is a reproduction-and-extension study of Leader Reward, not a new
architecture or inference search method.

\section{What changes relative to Leader Reward}

\subsection{Leader Reward's update}

For one instance, let $R_1,\ldots,R_n$ be rewards from the $n$ POMO starts,
$\bar R=n^{-1}\sum_iR_i$, $a_i=R_i-\bar R$, and
$\ell=\arg\max_iR_i$. With the main-phase multiplier $\alpha=40$, the
effective trajectory weights in \citet{wang2024leaderreward} are
\begin{equation}
  w_i^{\mathrm{LR}}
  =
  \begin{cases}
    a_i/\alpha, & i\neq\ell,\\
    a_\ell, & i=\ell.
  \end{cases}
  \label{eq:leader}
\end{equation}
Our implementation uses an additive leader coefficient $39$ and divides every
weight by $40$, which is algebraically identical. In the final two training
phases, only the leader receives a nonzero weight.

\subsection{The tested budget-indexed extension}

Our arm retains $n=100$ POMO starts and fixes $K=8$. Sort an instance's
rewards as $R_{(1)}\leq\cdots\leq R_{(n)}$. Before stabilization, the
implemented rank-gap weight is
\begin{equation}
  u_{(i)}
  =
  \begin{cases}
    0, & i<K,\\[1.5mm]
    \displaystyle
    \binom{n}{K}^{-1}
    \sum_{m=K}^{i}
    \binom{m-2}{K-2}
    \bigl(R_{(i)}-R_{(m-1)}\bigr), & i\geq K .
  \end{cases}
  \label{eq:rankgap}
\end{equation}
This distributes credit by rank and budget rather than assigning a special
role only to the current leader.

Two transformations define the recipe actually tested. Let
$\delta=0.01(R_{(n)}-R_{(1)})$. For $i\geq K$, the summed gap in
Equation~\eqref{eq:rankgap} is floored by replacing it with the larger of its
observed value and
$\delta\sum_{m=K}^{i}\binom{m-2}{K-2}$. The resulting vector
$\tilde{\boldsymbol u}$ is standardized within each instance:
\begin{equation}
  w_i^{\mathrm{stab}}
  =
  \frac{\tilde u_i-\overline{\tilde u}}
       {s(\tilde{\boldsymbol u})+10^{-8}},
  \qquad
  \mathcal L_{\mathrm{stab}}
  =
  -\sum_{i=1}^{n}
  \operatorname{stopgrad}(w_i^{\mathrm{stab}})
  \log\pi_\theta(\tau_i).
  \label{eq:stabilized}
\end{equation}
The floor and standardization alter the original score weights. ``Stabilized
Best-of-$K$'' therefore names an engineering recipe, not an unbiased Max@$K$
estimator. POMO also enumerates distinct starting nodes during training,
whereas Equation~\eqref{eq:bestofk} and the headline evaluation use independent
single-trajectory samples. The implemented update is a Best-of-$K$-motivated
surrogate, not a proof of the exact gradient of Equation~\eqref{eq:bestofk}.

\subsection{Context and attribution}

Best-of-$K$ and order-statistic policy optimization are established ideas.
PKPO \citep{pkpo2025} gives pass@$K$ and continuous-reward Max@$K$ estimators
with variance reduction; related formulations appear in MaxPO
\citep{maxpo2026} and OrderGrad \citep{ordergrad2026}. Poppy
\citep{grinsztajn2023poppy} uses a winner-takes-all objective across a
population of policies, a different architectural setting. Our contribution
is the controlled Leader Reward extension and its observed dependence on the
deployment budget, not the generic objective or estimator.

\section{Evaluation card}

\begin{center}
\small
\rowcolors{2}{white}{notepanel}
\begin{tabularx}{\linewidth}{>{\bfseries\color{noteink}}p{0.23\linewidth}Y}
\toprule
Field & Frozen or reported setting \\
\midrule
Problem and test set &
Euclidean TSP-100; 10,000 instances generated with seed 1234; reported mean
Concorde reference $7.765$ \citep{wang2024leaderreward}. \\
Training unit &
RL4CO POMO, batch size 64, 100 starts, 100,000 generated instances per epoch,
paired training seeds $\{0,1,2\}$. \\
Matched schedule &
3,050 epochs per arm: 2,900 at $10^{-4}$, 100 at
$5.5{\times}10^{-5}$, and 50 at $5.5{\times}10^{-6}$. \\
Primary readout &
Realized Best-of-8 cost from independent single-trajectory sampling;
same decoded-candidate count for both methods. \\
Secondary readouts &
Realized $K\in\{1,2,4,8,16,32,64,128\}$, one greedy trajectory, and
100-start $\times$ 8-augmentation greedy decoding. \\
Replication &
Training seed is the independent replication unit. Three paired seeds support
estimation only; the versioned design refuses confirmatory testing below six. \\
Evaluation runtime &
2,048-draw evaluation: 3,214--3,269 seconds per Leader Reward checkpoint and
3,227--3,246 seconds per stabilized checkpoint. \\
Unavailable telemetry &
Training wall-clock, energy, peak memory, and per-step NaN/clamp counters were
not frozen. Their equivalence or absence is not claimed. \\
\bottomrule
\end{tabularx}
\end{center}

For each checkpoint and test instance, evaluation draws a fresh pool of 2,048
independent trajectories. Separately at each $K$, the pool is partitioned into
disjoint size-$K$ blocks; the best cost in each block is averaged. The
augmented-greedy endpoint instead considers 800 structured candidates and is
reported as a reproduction check, not an equal-budget comparison with sampled
Best-of-$K$.

The fixed instances and within-instance blocks reduce evaluation noise but are
not extra training replications. We report all three paired rows, their mean,
and a descriptive seed-level BCa interval. With three nonzero pairs, the
smallest attainable one-sided signed-rank $p$-value is $1/8$. All reported
outputs are finite and no defensive weight clamp was configured; the missing
training telemetry prevents a stronger numerical-stability statement.

\section{Finding 1 --- the Leader Reward endpoint reproduces}

Under 100-start, 8-augmentation greedy decoding, the three-seed Leader Reward
mean is $7.7662$, matching the published $7.766$ at its displayed precision.
This is a direct check on the baseline and test protocol. The stabilized arm
obtains $7.7669$ and is slightly worse in this regime.

\section{Finding 2 --- sampled Best-of-8 improves in all three seeds}

The stabilized arm has lower sampled Best-of-8 cost in every paired seed
(Table~\ref{tab:paired}). Its seed mean is $7.7944$, compared with $7.8136$
for Leader Reward. The paired difference, Leader Reward minus stabilized, is
$+0.0193$, with a descriptive BCa interval of $[+0.0068,+0.0257]$. This is a
$0.247\%$ cost reduction relative to Leader Reward. Relative to the common
$7.765$ Concorde reference, the remaining gap changes from $0.626\%$ to
$0.378\%$, a $39.7\%$ reduction. None of these summaries is a confirmatory
significance result.

\begin{table}[!htbp]
\centering
\caption{Paired realized Best-of-8 sampling cost. Positive differences favor
the stabilized recipe.}
\label{tab:paired}
\begin{tabular}{rrrr}
\toprule
Training seed & Leader Reward & Stabilized Best-of-$K$ & Difference \\
\midrule
0 & 7.8192 & 7.7933 & +0.0259 \\
1 & 7.8094 & 7.8027 & +0.0068 \\
2 & 7.8122 & 7.7870 & +0.0252 \\
\midrule
Mean & 7.8136 & 7.7944 & +0.0193 \\
\bottomrule
\end{tabular}
\end{table}

\section{Finding 3 --- the advantage depends on sampling budget}

\begin{figure}[!htbp]
  \centering
  \includegraphics[width=0.74\linewidth]{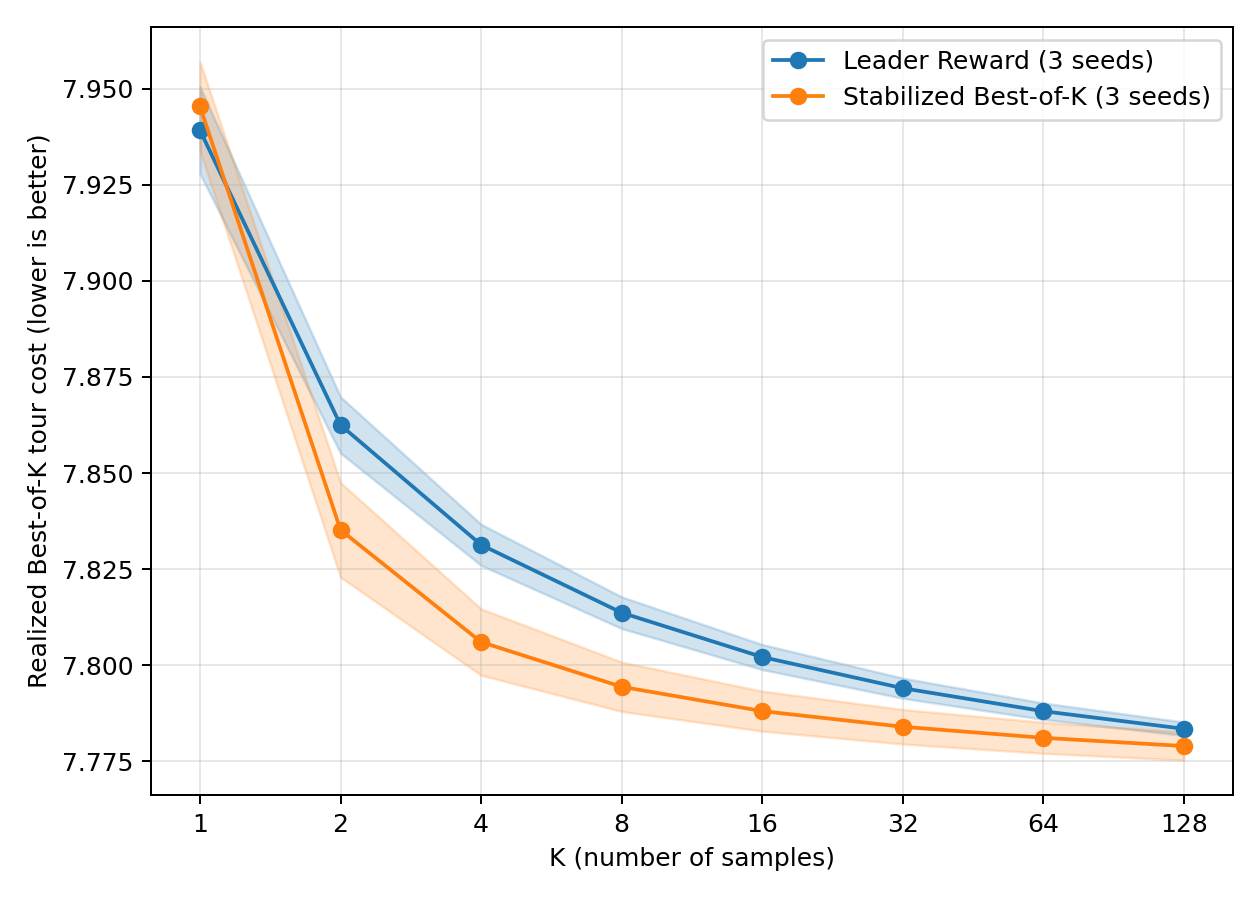}
  \caption{Realized sampled Best-of-$K$ TSP-100 cost. Points are means over
  three training seeds; bands show one population standard deviation across
  seeds. Lower is better.}
  \label{fig:curve}
\end{figure}

The seed mean favors the stabilized arm from $K=2$ through $K=128$
(Figure~\ref{fig:curve} and Table~\ref{tab:curve}), but the advantage contracts
as $K$ grows. Leader Reward is better at $K=1$. At $K=128$, the stabilized
mean remains lower while seed 1 reverses by $0.0011$. These secondary results
describe the curve; they are not separate confirmatory tests.

\begin{table}[!htbp]
\centering
\caption{Seed-mean realized Best-of-$K$ sampling cost. Lower is better.}
\label{tab:curve}
\small
\resizebox{\linewidth}{!}{%
\begin{tabular}{lcccccccc}
\toprule
Method & $K=1$ & $K=2$ & $K=4$ & $K=8$ & $K=16$ & $K=32$ & $K=64$ & $K=128$ \\
\midrule
Leader Reward
& \textbf{7.9392} & 7.8625 & 7.8313 & 7.8136 & 7.8021 & 7.7940 & 7.7880 & 7.7834 \\
Stabilized Best-of-$K$
& 7.9454 & \textbf{7.8352} & \textbf{7.8060} & \textbf{7.7944}
& \textbf{7.7880} & \textbf{7.7839} & \textbf{7.7811} & \textbf{7.7789} \\
\bottomrule
\end{tabular}}
\end{table}

\section{Finding 4 --- the gain does not transfer to every decoder}

The stabilized policy improves one greedy trajectory and sampled Best-of-8,
but Leader Reward is lower for sampled Best-of-1 and the original
augmented-greedy protocol (Table~\ref{tab:decoders}). Under augmented greedy,
Leader Reward is better in all three seeds by $0.0005$--$0.0009$. Candidate
budgets differ across rows; the rows are distinct deployment questions rather
than equal-compute comparisons.

\begin{table}[!htbp]
\centering
\caption{Seed-mean TSP-100 cost under distinct deployment protocols.}
\label{tab:decoders}
\small
\begin{tabular}{lrr}
\toprule
Deployment protocol & Leader Reward & Stabilized Best-of-$K$ \\
\midrule
One sampled trajectory & \textbf{7.9392} & 7.9454 \\
One greedy trajectory & 7.8948 & \textbf{7.8664} \\
Best of 8 sampled trajectories & 7.8136 & \textbf{7.7944} \\
Best of 128 sampled trajectories & 7.7834 & \textbf{7.7789} \\
100 starts $\times$ 8 augmented greedy & \textbf{7.7662} & 7.7669 \\
\bottomrule
\end{tabular}
\end{table}

\section{What this evidence does not establish}

\begin{summarybox}
\notelabel{Claim boundary}

\vspace{0.35em}
\begin{tabularx}{\linewidth}{@{}>{\bfseries\color{noteink}}p{0.27\linewidth}Y@{}}
Statistical confirmation &
Three training seeds establish a repeated observation, not the planned
six-seed confirmatory result. The BCa interval is fragile at this sample size,
and the experiment was not externally preregistered. \\[2pt]
Generality &
The evidence covers one POMO implementation and Euclidean TSP-100. It does not
cover CVRP, FFSP, other POMO families, or out-of-distribution instances. \\[2pt]
Estimator identity &
Gap flooring, standardization, and the distinct-start training law prevent an
unbiased-gradient interpretation. The experiment does not isolate which
stabilizer causes the observed change. \\[2pt]
Comparator coverage &
The campaign has no newly trained vanilla POMO, objective-faithful Max@$K$ arm,
or component ablations. \\[2pt]
Universal superiority &
Leader Reward remains better at sampled $K=1$ and under its original
augmented-greedy protocol. No NCO state-of-the-art claim is made.
\end{tabularx}
\end{summarybox}

\section{Artifacts and reproducibility}

The arXiv source package includes the complete per-seed sampling curve,
greedy and augmented-greedy rows, evaluation runtimes, the figure-generation
script, and an artifact manifest. Frozen repository artifacts record the
checkpoints, configuration, hashes, and per-instance evaluation outputs.

\begin{summarybox}
\notelabel{Bottom line}

\vspace{0.35em}
On this three-seed TSP-100 study, a stabilized $K=8$ rank-gap signal is a
promising extension of Leader Reward when deployment samples a modest number
of independent candidates. The result is specific to that deployment regime
and is not evidence of a generally better POMO policy.
\end{summarybox}

\begingroup
\small
\setlength{\bibsep}{2pt}
\bibliographystyle{plainnat}
\bibliography{references}
\endgroup

\end{document}